\documentclass[journal,twoside,web]{ieeecolor}
\usepackage{generic}
\usepackage{cite}
\usepackage{amsmath,amssymb,amsfonts}
\usepackage{graphicx}
\usepackage{textcomp}
\usepackage{bm}
\usepackage{xcolor} 
\usepackage{threeparttable}
\usepackage{booktabs}
\usepackage{multirow}
\usepackage{hyperref}
\usepackage{pifont}
\usepackage{color}
\usepackage{soul}
\usepackage{colortbl}

\def\BibTeX{{\rm B\kern-.05em{\sc i\kern-.025em b}\kern-.08em
    T\kern-.1667em\lower.7ex\hbox{E}\kern-.125emX}}
\author{Yunkang Cao,
Haiming Yao,
Wei Luo, 
Weiming Shen,~\IEEEmembership{Fellow,~IEEE}
\thanks{Manuscript received XX XX, 2024; revised XX XX, 20XX. This work was supported in part by Ministry of Industry and Information Technology of the People's Republic of China under Grant \#2023ZY01089. The computation is completed in the HPC Platform of Huazhong University of Science and Technology. (\textit{Corresponding author: Weiming Shen}.) }
\thanks{Yunkang Cao and Weiming Shen are with the State Key Laboratory of Intelligent Manufacturing Equipment and Technology, Huazhong University of Science and Technology, Wuhan 430074, China (e-mail:  cyk\_hust@hust.edu.cn; wshen@ieee.org).} 
\thanks{Haiming Yao and Wei Luo are with the State Key Laboratory of Precision Measurement Technology and Instruments, Department of Precision Instrument, Tsinghua University, Beijing 100084, China. (e-mails: yhm22@mails.tsinghua.edu.cn; luow23@@mails.tsinghua.edu.cn).}
}

\begin{document}

\title{VarAD: Lightweight High-Resolution Image Anomaly Detection via Visual \\Autoregressive Modeling}

\newcommand{\cyk}[1]{\textcolor{red}{cyk: #1}}
\newcommand{\ie}{\textit{i.e.}}
\newcommand{\etc}{\textit{etc.}}
\newcommand{\eg}{\textit{e.g.}}

\maketitle

\begin{abstract}

This paper addresses a practical task: High-Resolution Image Anomaly Detection (HRIAD). In comparison to conventional image anomaly detection for low-resolution images, HRIAD imposes a heavier computational burden and necessitates superior global information capture capacity. To tackle HRIAD, this paper translates image anomaly detection into visual token prediction and proposes VarAD based on visual autoregressive modeling for token prediction. Specifically, VarAD first extracts multi-hierarchy and multi-directional visual token sequences, and then employs an advanced model, Mamba, for visual autoregressive modeling and token prediction. During the prediction process, VarAD effectively exploits information from all preceding tokens to predict the target token. Finally, the discrepancies between predicted tokens and original tokens are utilized to score anomalies. Comprehensive experiments on four publicly available datasets and a real-world button inspection dataset demonstrate that the proposed VarAD achieves superior high-resolution image anomaly detection performance while maintaining lightweight, rendering VarAD a viable solution for HRIAD. Code is available at \href{https://github.com/caoyunkang/VarAD}{\url{https://github.com/caoyunkang/VarAD}}.

\end{abstract}


\section{Introduction}
\label{sec:introduction}

\IEEEPARstart{I}{mage} anomaly detection (AD) aims to identify irregular patterns within images, playing a crucial role in industrial defect inspection~\cite{IM-IAD}. While existing AD methods~\cite{CDO,Patchcore} excel in low-resolution settings, typically $256 \times 256$, and achieve high detection performance on standard datasets like MVTec AD~\cite{MVTec-AD}, real-world applications often demand high-resolution images ($1024 \times 1024$ or higher) for inspection.

This requirement is due to the potential defects being extremely minute compared to the overall product, rendering low-resolution AD methods less effective~\cite{STCIKD}. Although it is feasible to directly downsample high-resolution images to a lower resolution for inspection, this process can cause the loss of critical details regarding subtle anomalies. For instance, certain scenarios like plastic part inspection~\cite{CDO} and lens inspection~\cite{vision-datasets} necessitate detecting anomalies of approximately 1mm$\times$1mm in a product measuring 100mm$\times$100mm. Such anomalies are incredibly challenging to detect in low-resolution ($256 \times 256$) images, as they occupy only about five pixels. Conversely, in high-resolution images ($1024 \times 1024$), these anomalies occupy around 100 pixels, significantly easing the detection process. While it can be feasible to apply a sliding window strategy~\cite{dong2023swssl} to high-resolution images to obtain smaller patches instead of downsampling, this approach can lose global information within images and lead to unreliable detection results. Hence, this study proposes to detect anomalies directly in high-resolution images, namely High-Resolution Image Anomaly Detection (HRIAD).

\begin{figure}[t]
    \centering
    \includegraphics[width=1\linewidth]{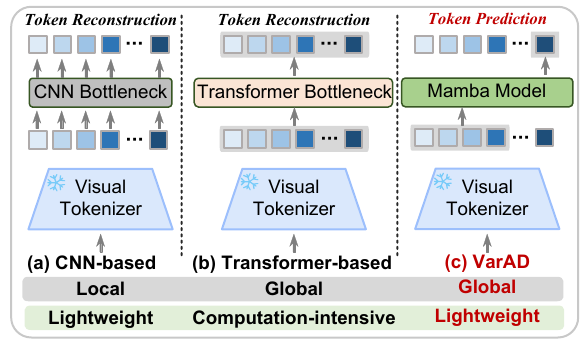}
    \caption{\textbf{Motivation.} By translating anomaly detection into token prediction, VarAD demonstrates the ability to capture global information while remaining lightweight.
    }
    \label{fig:teaser}
    \vspace{-5mm}
\end{figure}

In comparison to the typical low-resolution AD setting, HRIAD faces several inherent challenges. Thus, the computational burden increases significantly with high-resolution images, and capturing global information, which can be critical for detecting certain anomalies, is more challenging compared to low-resolution counterparts.
% Existing AD methods, such as RD4AD~\cite{RD4AD}, primarily rely on local image embeddings (tokens) extracted from pre-trained Convolutional Neural Networks (CNNs) to detect anomalies. However, these tokens typically exhibit a limited receptive field and contain minimal global information. This deficiency becomes critical in AD tasks where comprehensive scene understanding is essential.
Some reconstruction-based AD methods, like RD4AD~\cite{RD4AD}, can capture a certain level of global information. Specifically, these methods begin by extracting visual tokens using a vision tokenizer and then reconstructing these tokens through a bottleneck module. Anomaly scores are derived from the discrepancies between the original and reconstructed visual tokens. The bottleneck can be based on Convolutional Neural Networks (CNNs) or vision transformers. However, while the CNN-based bottleneck~\cite{RD4AD} (Fig.~\ref{fig:teaser} (a)) contributes to capturing neighboring information for reconstruction, its receptive fields remain limited and cannot utilize all tokens for reconstruction, potentially diminishing their effectiveness for HRIAD. On the other hand, transformer-based bottlenecks~\cite{PSA-VT,Intra} (Fig.~\ref{fig:teaser} (b)) can achieve global receptive fields through self-attention. However, the self-attention mechanism leads to a significant increase in computational demands due to the quadratic complexity of attention calculations, making transformer-based bottlenecks impractical for the HRIAD task.

Therefore, this study aims to introduce a lightweight HRIAD model capable of capturing global information. Specifically, the study proposes VarAD, which employs Visual AutoRegressive (VAR) modeling to capture the sequential relationships between visual tokens. VarAD draws inspiration from the success of autoregressive models in large language models (LLMs)\cite{openai2024gpt4}, renowned for their robust global modeling capacity through next-token prediction. Similarly, VarAD formulates the HRIAD task as a token prediction task, as illustrated in Fig.\ref{fig:teaser} (c). It initially tokenizes images into a sequence and trains the model to predict future tokens based on previous ones. By training on normal images, the model is expected to predict only normal tokens during testing. Discrepancies between predicted and original tokens are then utilized to score anomalies. In contrast to reconstruction-based approaches with transformer-based bottlenecks~\cite{AMI-Net,PNPT,GLCF}, which employ all input tokens to reconstruct normal tokens, our method uses only preceding tokens to predict future ones. With advanced visual autoregressive models, we reduce complexity to a linear scale while preserving global information capture capabilities.

To enhance detection performance, VarAD proposes to adapt a pre-trained vision model (DINO~\cite{dinov2}) as the vision tokenizer for extracting visual tokens. These multi-hierarchy visual tokens are then traversed spatially via multiple directions into multiple token sequences. These multi-hierarchy and multi-directional visual token sequences provide informative contexts and enable comprehensive prediction of normal tokens. Furthermore, VarAD leverages Mamba~\cite{mamba}, an up-to-date autoregressive model for token prediction, achieving superior modeling capacity and efficiency. Experimental results on four widely used public AD datasets demonstrate the effectiveness and efficacy of VarAD.

In summary, this study makes the following contributions:

\begin{itemize}
    \item This study addresses a more practical setting, \ie, high-resolution image anomaly detection, and conducts systematic benchmarks on this setting, making a step towards practical industrial applications. 
    While previous studies~\cite{STCIKD} have introduced HRIAD, our benchmark is more comprehensive, encompassing seven methods across four datasets with resolutions ranging from $256\times256$ to $1024\times1024$.
    \item This study proposes to reformulate image anomaly detection as a token prediction task. VarAD based on visual autoregressive modeling is proposed to address this task. 
    \item VarAD proposes to extract multi-hierarchy and multi-directional visual token sequences and utilizes Mamba for autoregressing, achieving better detection performance. 
\end{itemize}

\begin{figure*}[t]
    \centering
    \includegraphics[width=.8\linewidth]{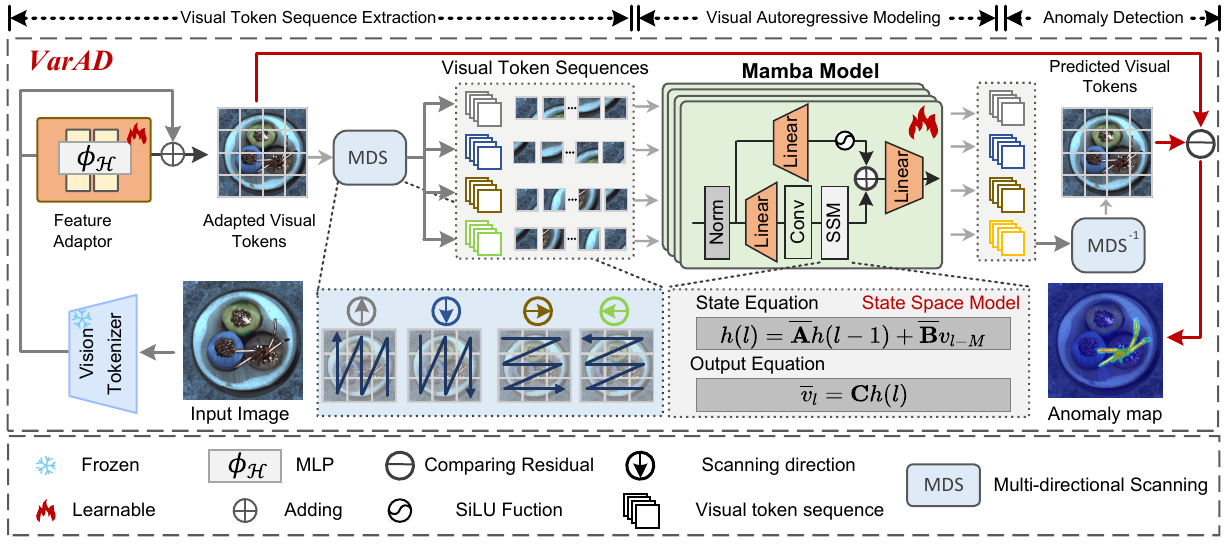}
    \caption{\textbf{Framework of the proposed Visual AutoRegressive modeling-based Anomaly Detection (VarAD).}}
    \label{fig:framework}
\end{figure*}

\section{Related Work}

\subsection{Image Anomaly Detection}
This study classifies existing image anomaly detection methods based on the type of information utilized, namely, local and global information.

\subsubsection{Local Information}
Many image anomaly detection methods focus on extracting local information. These methods typically employ pre-trained CNNs to extract image embeddings, which are then used to model the distribution of these embeddings. Anomalies are detected based on discrepancies between testing and training embeddings. Existing methods can be categorized into knowledge distillation~\cite{DAF,PFM,BiaS}, memory bank~\cite{Patchcore,GCPF}, discriminative~\cite{GLASS,Simplenet}, 
and flow-based~\cite{PyramidFlow,MSFlow} methods. For example, knowledge distillation-based methods~\cite{DAF} align a pre-trained teacher network with a randomly initialized student network via normal images. They assume that since the student and teacher networks are only aligned via normal images, they will have different outputs for abnormal images, and these discrepancies are utilized to detect anomalies. Memory bank-based methods~\cite{Patchcore,GCPF}, on the other hand, store representative normal embeddings and score anomalies based on the nearest distance to this bank. 
SWSSL~\cite{dong2023swssl} introduces self-supervised learning for augmentation-invariant features and combines these features with memory banks for high-resolution medical image anomaly detection. However, SWSSL still relies on a patch encoder for feature extraction, which can only perceive local information.
{Discriminative approaches like SimpleNet~\cite{Simplenet} and GLASS~\cite{GLASS} typically train a network to differentiate between normal and abnormal features for anomaly detection; however, such networks are often implemented using simple convolutional layers, which are limited to capturing local information. }
In addition, flow-based methods~\cite{PyramidFlow,MSFlow} use flow models to model normal embedding distribution directly. While these methods have achieved promising detection performance for traditional low-resolution scenarios, their effectiveness may diminish in high-resolution images due to their limited capacity to capture global information.

\subsubsection{Global Information}
Certain image anomaly detection methods enhance their capacity to capture global information by modifying their model architectures. Conventional image anomaly detection methods are based on CNNs. For instance, CDO~\cite{CDO} replaces conventional backbones with HRNet~\cite{HRNet} to extract image embeddings with larger receptive fields. RD4AD~\cite{RD4AD} and OCR-GAN~\cite{OCR-GAN} introduce bottlenecks to extract global context, followed by embedding reconstruction based on global semantic information. Additionally, methods like GLCF~\cite{GLCF} establish two network branches (global and local) to extract and aggregate global and local embeddings, respectively, thereby improving the capture of global information. {Similarly, EfficientAD~\cite{EfficientAD} extends the knowledge-distillation scheme by operating in local and global feature spaces simultaneously.  }
Some approaches explicitly target the capture of global information by using visual transformers for reconstruction, such as PSA-VT~\cite{PSA-VT}, PNPT~\cite{PNPT}, and MLDFR~\cite{MLDFR}, or for inpainting, like Intra~\cite{Intra} and AMI-Net~\cite{AMI-Net}.
However, the attention mechanisms in transformers often lead to intensive computational demands due to the quadratic complexity of attention calculations in both computation and memory, hindering their applications in HRIAD. In contrast, STCIKD~\cite{STCIKD} adopts a strategy to predict future tokens based on several preceding tokens but still fails to extract information from all preceding tokens. To capture global information and enable lightweight detection, the proposed VarAD also translates the anomaly detection task into token prediction and exploits visual autoregressive modeling for great efficiency.

\subsection{Autoregressive Modeling}

\subsubsection{Language Autoregressive Modeling}
Language autoregressive modeling has gained significant traction for its promising global capture capacity, exemplified by GPT-4~\cite{openai2024gpt4}. A recent advancement in this field is Mamba~\cite{mamba}, which allows each token in a sequence to interact with previously scanned tokens via a compressed hidden state, reducing quadratic complexity to linear. The efficacy of Mamba in long sequence modeling has garnered considerable attention.

\subsubsection{Visual Autoregressive Modeling}
The success of language autoregressive modeling has spurred efforts in visual autoregressive modeling. Early work, such as VQGAN~\cite{VQ-GAN}, employs a decoder-only transformer to generate image tokens in a standard raster-scan autoregressive manner, leading to improved image generation quality. More recently, visual autoregressive modeling has been utilized to construct visual foundation models. For instance, the Large Vision Model (LVM)~\cite{LVM} translates visual data into visual sentences and utilizes autoregressive training for next token prediction, enabling the solution of various vision tasks by designing suitable visual prompts at test time. The success of Mamba has also inspired numerous works in the field of visual data. For example, the pioneering endeavor VMamba~\cite{vmamba} proposes traversing the spatial domain, converting images into multi-directional visual token sequences, and then utilizing Mamba for modeling, showcasing promising capabilities across various visual perception tasks. Drawing inspiration from these advancements in visual autoregressive modeling, this study chooses to reformulate anomaly detection as a token prediction task and utilizes Mamba for visual autoregressive modeling. The proposed VarAD presents to have better HRIAD performance and higher efficiency.

\section{VarAD methodology}

\subsection{Problem Definition}
The HRIAD task is designed to meet practical requirements. Following the mainstream unsupervised anomaly detection setting~\cite{CDO}, the training set $\mathcal{T}_{\text{train}}$ for HRIAD consists of a collection of high-resolution images ($1024\times1024$ in this study) representing normal instances of a specific product category. The objective is to develop a model capable of identifying anomalies within unseen images from the same category and generating a corresponding anomaly map $\mathbf{A} \in \mathbb{R}^{H\times W}$, where $H$ and $W$ denote the resolution of the original images. In the anomaly map, higher values indicate that the corresponding coordinate position is more likely to be abnormal. Typically, the highest score of the anomaly map can indicate the overall anomaly degree of the whole image.

\subsection{Method Overview}
The proposed HRIAD method, VarAD, introduces a novel approach by reformulating anomaly detection as a token prediction task. As shown in Fig.~\ref{fig:framework}, this method comprises several key steps. Firstly, given an image, VarAD extracts multi-hierarchy and multi-directional visual token sequences. Subsequently, in the visual autoregressive modeling phase, VarAD leverages Mamba~\cite{mamba} to predict future tokens based on previous tokens in an autoregressive manner. Following this, VarAD computes the prediction errors between the predicted tokens and the original tokens to assess anomalies. Finally, the results from multi-hierarchy and multi-directional predictions are aggregated to produce the final anomaly detection results.

\subsection{Visual Token Sequence Extraction}
The visual token sequence extraction step consists of two sub-steps: tokenizing the image into multi-hierarchy visual tokens, which are then sequentialized into multi-directional visual token sequences.

\subsubsection{Image Tokenized} 
The descriptive quality of visual tokens, as described in LVM~\cite{LVM}, plays a crucial role in the autoregressive modeling process. VarAD proposes to adapt a pre-trained vision model, DINO~\cite{dinov2}, as the visual tokenizer. Specifically, the image is inputted into the pre-trained visual model, and tokens from specific layers are extracted to form multi-hierarchy visual tokens $\mathbf{F}^{h} \in \mathbb{R}^{C \times \frac{H}{N^{h}} \times \frac{W}{N^{h}}}$, where $N^{h}$ is the downsampling ratio of $h$-th hierarchy and $\mathcal{H}$ is the number of hierarchies. However, since the pre-trained model is trained on natural image datasets, its descriptiveness may diminish when applied to the target category due to domain gap issues. Therefore, VarAD addresses this challenge by adapting the pre-trained DINO model. This adaptation involves integrating a feature adapter $\phi_{h}(\cdot)$ for each hierarchy feature and utilizing a residual connection to obtain the adapted tokens $\mathbf{\hat{F}}^{h}$:

\begin{equation}
\mathbf{\hat{F}}^{h} = \phi_{h}(\mathbf{F}^{h}) + \mathbf{F}^{h}, \quad h=1,\ldots,\mathcal{H}
\end{equation}

\noindent 
% where the adapter $\phi_{h}(\cdot)$ is implemented as a simple linear layer in this study. 

During training, the visual model remains frozen, while the adapters are trainable. This approach mitigates the domain gap without compromising the original descriptiveness.

\subsubsection{Token Sequentialized}
Autoregressive modeling is naturally designed for 1-D sequences and cannot be directly applied to 2-D visual tokens. Hence, VarAD employs a spatial traversal of the visual tokens and translates them into visual token sequences. A straightforward strategy might involve expanding the visual tokens linearly along the row axis. However, this approach restricts the token prediction process to only consider preceding tokens, thereby potentially sacrificing valuable post-target information. To overcome this limitation, VarAD introduces a multi-directional scanning function denoted as $\text{MDS}$, which scans visual tokens in multiple directions. This innovation allows for the comprehensive retention of information crucial for accurate token prediction. Illustrated in Fig.~\ref{fig:framework}, VarAD unfolds visual tokens along rows and columns into sequences, proceeding to scan along four distinct directions: top-left to bottom-right, bottom-right to top-left, top-right to bottom-left, and bottom-left to top-right. This ensures that each token integrates information from all tokens, enhancing predictive accuracy across various directions. The process is formulated as follows:

\begin{equation}
\{v_1,\ldots,v_{L^{h}}\}^{h}_{k} = \text{MDS}(\mathbf{\hat{F}}^{h}), \quad k=1,2,3,4
\end{equation}

\noindent where $\{v_1,\ldots,v_{L^{h}}\}^{h}_{k}$ represents the $k$-th directional visual token sequence of the $h$-th hierarchy. $L^{h}$ denotes the length of the visual token sequences, which is equal to $\frac{H}{N^{h}} \times \frac{W}{N^{h}}$.

\subsection{Visual Autogressive Modeling}

VarAD utilizes visual autoregressive modeling for token prediction. Specifically, Mamba~\cite{mamba} is utilized as the autoregressive model.

\subsubsection{Mamba}
The architecture of the Mamba model is visualized in Fig.~\ref{fig:framework}, within which the state space model (SSM) is the most vital component. Specifically, SSMs are conventionally recognized as linear time-invariant systems mapping stimulation $x(t) \in \mathbb{R}^L$ to response $y(t) \in \mathbb{R}^L$. These SSMs are typically formulated as linear ordinary differential equations (ODEs) (Eq.~\eqref{ODEs}), with parameters including $\mathbf{A} \in \mathbb{R}^{P \times P}$, $\mathbf{B}, \mathbf{C} \in \mathbb{R}^{P}$ for a state size $P$.

\begin{equation}\label{ODEs}
    h'(t) = \mathbf{A}h(t) + \mathbf{B}x(t), \quad
    y(t) = \mathbf{C}h(t)
\end{equation}

The continuous parameters $\mathbf{A}$ and $\mathbf{B}$ can be discretized from the continuous system into discrete parameters $\overline{\mathbf{A}}$ and $\overline{\mathbf{B}}$ with zero-order hold with a timescale parameter $\triangle$:

\begin{equation}\label{discretization}
\begin{aligned}
    \overline{\mathbf{A}} &= \exp(\triangle{\mathbf{A}}), \\
    \overline{\mathbf{B}} &= (\triangle{\mathbf{A}})^{-1}(\exp(\triangle{\mathbf{A}})-\mathbf{I}) \cdot \triangle{\mathbf{B}}.
\end{aligned}
\end{equation}

Post discretization, the model can be represented as:

\begin{equation}\label{discretized_ODEs}
    h(t) = \overline{\mathbf{A}}h(t-1) + \overline{\mathbf{B}}x(t), \quad
    y(t) = \mathbf{C}h(t)
\end{equation}

\noindent In addition, Mamba associates the matrices $\mathbf{A}, \mathbf{B}, \mathbf{C}$ with the input, thereby ensuring the dynamism of weights within autoregressive modeling. 

\subsubsection{Token Prediction}
VarAD utilizes Mamba to predict the $l$-th token based on the previous tokens. However, due to the potentially strong correlation between neighboring visual tokens, the conventional autoregressive approach used in language modeling, \ie, next-token prediction, tends to over-focus on nearby tokens for visual data. This over-emphasis on neighboring tokens may hinder the learning of global information. Therefore, VarAD proposes to predict the $l$-th token based on previous tokens, excluding the $M$ nearest tokens, denoted by $\{v_1,\ldots,v_{l-M}\}$. Given that the resolutions of features vary across hierarchies, this study associates $M$ with the feature resolutions, setting $M^{h}=m\frac{H}{N^{h}}$ and $M^{h}=m\frac{W}{N^{h}}$ for the row-spanned and column-spanned token sequences, respectively, where $m$ is a hyperparameter named prediction step that controls the length of tokens to be excluded.

Then, Eq.~\eqref{discretized_ODEs} for predicting the $l$-th token is rewritten as:

\begin{equation}\label{one_token_prediction}
    h(l) = \overline{\mathbf{A}}h(l-1) + \overline{\mathbf{B}}v_{l-M}, \quad
    \overline{v}_{l} = \mathbf{C}h(l)
\end{equation}

\noindent where $h(l)$ denotes the hidden state at the $l$-th step, incorporating information from $\{v_1,\ldots,v_{l-M}\}$, and $\overline{v}_{l}$ represents the predicted token. Thus, the token prediction process maintains linear complexity, ensuring high efficiency even for high-resolution images.

To predict the first $M$ tokens, VarAD adds [BOS] (beginning of sequence) tokens to the beginning of each visual token sequence and [EOS] (end of sequence) tokens to the end, where $[\text{BOS}],[\text{EOS}] \in \mathbb{R}^{M \times C}$. Then, utilizing Eq.~\eqref{one_token_prediction}, VarAD obtains the predicted tokens as follows:

\begin{equation}\label{all_token_prediction}
\{[\text{BOS}], v_1,\ldots,v_{L}\} \longrightarrow \{\overline{v}_1,\ldots,\overline{v}_{L}, [\text{EOS}]\}
\end{equation}

\begin{figure*}[t]
    \centering
    \includegraphics[width=.85\linewidth]{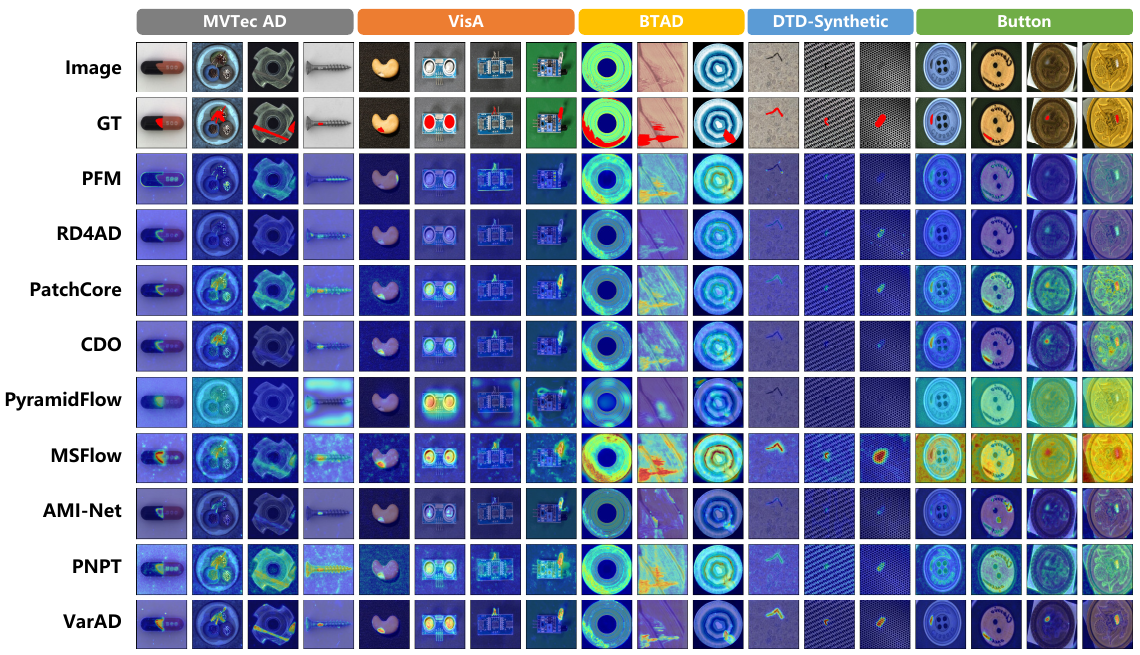}
        \caption{\textbf{Anomaly maps on the four publicly available datasets and our private data from the real-world button inspection task.} }
    \label{fig:main_visual}
\end{figure*}

\subsection{Anomaly Detection}

The hypothesis of VarAD is that through training solely with normal samples, the autoregressive model can exclusively predict normal tokens, and the disparities between predicted tokens and original tokens can be utilized to score anomalies. With Eq.~\eqref{all_token_prediction}, VarAD has obtained predicted tokens for multi-hierarchy and multi-directional visual token sequences. Subsequently, VarAD proposes to aggregate predicted tokens from multiple directions, thus ensuring that the prediction encompasses information from all tokens (excluding neighboring ones). The process is formulated as:

\begin{equation}
\mathbf{\overline{F}}^{h} = \text{MDS}^{-1}(\{\overline{v}_1,\ldots,\overline{v}_{L^{h}}\}^{h}_{k}), \quad k=1,2,3,4
\end{equation}

\noindent where $\text{MDS}^{-1}$ represents the reverse function of the multi-directional scanning process $\text{MDS}$, which first traverses individual 1-D predicted token sequences back to 2-D visual tokens as like $\mathbf{\hat{F}}^{h}$, and then averages all the predicted 2-D tokens into the predicted visual tokens $\mathbf{\overline{F}}^{h}$.

Subsequently, the anomaly maps $\mathbf{A}$ by summarizing the prediction errors across multiple hierarchies:

\begin{equation}
\begin{aligned}
\mathbf{A}^{h}_{ij} &= ||\mathbf{\overline{F}}^{h}_{ij}-\mathbf{\hat{F}}^{h}_{ij}||^2_2, \\
\mathbf{A}_{ij} &= \sum_{h=1}^{\mathcal{H}}(\mathbf{A}^{h}_{ij})
\end{aligned}
\end{equation}

During training, VarAD minimizes the anomaly maps $\mathbf{A}$ for normal images, thereby ensuring that the predicted tokens are similar to the original tokens, formulated as:

\begin{equation}
\begin{aligned}
\mathcal{L}=\sum_{ij}{\mathbf{A}_{ij}}
\end{aligned}
\end{equation}

\noindent By optimizing VarAD with normal images through this objective, the prediction errors for normal tokens are expected to be minimized, while the errors for abnormal tokens can be larger, as VarAD is not trained to predict abnormal tokens.

\section{Experiments}

\subsection{Experiments Setup}
\subsubsection{Dataset Descriptions}
This study systematically evaluates the proposed VarAD on four publicly available datasets: MVTec AD \cite{MVTec-AD}, VisA \cite{VisA}, BTAD \cite{BTAD}, and DTD-Synthetic \cite{DTD}. The statistical information of these datasets is summarized in Table~\ref{tab:dataset_info}, encompassing 42 categories in total, with 15287 normal training samples, and 2237 and 3695 normal and abnormal testing samples, respectively. These extensive datasets provide a comprehensive HRIAD benchmark.

% \usepackage{booktabs}
% \usepackage{multirow}
\begin{table}[h]
\centering
\caption{Statistical information of the utilized datasets. \# denotes the number of categories or samples.}
\label{tab:dataset_info}
\resizebox{1\linewidth}{!}{
\begin{tabular}{@{}cc|cccc|c@{}}
\toprule[1.5pt]
\multicolumn{2}{c|}{Dataset}           & MVTec AD & VisA & BTAD & DTD-Synthetic & Summary \\ \midrule
\multicolumn{2}{c|}{\#Category}       & 15       & 12   & 3    & 12            & 42      \\ 
Training                 & \#Normal   & 3629     & 8659 & 1799 & 1200          & 15287   \\
\multirow{2}{*}{Testing} & \#Normal   & 467      & 962  & 451  & 357           & 2237    \\
                         & \#Abnormal & 1258     & 1200 & 290  & 947           & 3695    \\ \bottomrule[1.5pt]
\end{tabular}
}
\end{table}

% \usepackage{booktabs}
\begin{table*}[t]
\centering
\caption{Quantitative comparisons of VarAD with alternative methods on public datasets. Results are presented as pixel-level AUROC\%/max-F1\%/AP\%. Best scores are highlighted in \textbf{bold}, while the second-best scores are also \underline{underlined}.}
\label{tab:main_pixel}
\resizebox{1.\linewidth}{!}{
{
\begin{tabular}{ccccccccc>{\columncolor{blue!8}}c}
\toprule[1.2pt]
Dataset &
\begin{tabular}[c]{@{}c@{}}PFM~\cite{PFM}\\      TII'2022\end{tabular} &
\begin{tabular}[c]{@{}c@{}}RD4AD~\cite{RD4AD}\\      CVPR'2022\end{tabular} &
\begin{tabular}[c]{@{}c@{}}PatchCore~\cite{Patchcore}\\      CVPR'2022\end{tabular} &
\begin{tabular}[c]{@{}c@{}}CDO~\cite{CDO}\\      TII'2023\end{tabular} &
\begin{tabular}[c]{@{}c@{}}PyramidFlow~\cite{PyramidFlow}\\      CVPR'2023\end{tabular} &
\begin{tabular}[c]
{@{}c@{}}MSFlow~\cite{MSFlow}\\      TNNLS'2024\end{tabular} &
\begin{tabular}[c]{@{}c@{}}AMI-Net~\cite{AMI-Net}\\      TASE'2024\end{tabular} &
\begin{tabular}[c]{@{}c@{}}PNPT~\cite{PNPT}\\      TII'2024\end{tabular} &
\begin{tabular}[c]{@{}c@{}}\textbf{VarAD}\\      \textbf{Proposed}\end{tabular} \\ \midrule
MVTec AD~\cite{MVTec-AD} &
  91.3/36.6/27.9 &
  93.4/49.6/42.5 &
  95.8/60.1/60.5 &
  96.4/\underline{62.8}/\underline{62.9} &
  94.2/48.9/43.4 &
  \underline{97.2}/62.4/62.7 &
  94.9/53.2/49.9 &
  94.8/52.3/50.8 &
  \textbf{97.7}\textbf{/63.8}/\textbf{63.0} \\
VisA~\cite{VisA} &
  94.6/14.2/6.7 &
  92.9/39.0/26.8 &
  95.1/\underline{46.6}/38.9 &
  \underline{97.6}/46.5/37.7 &
  90.8/32.4/24.2 &
  96.9/43.9/37.4 &
  96.9/38.6/34.0 &
  96.7/45.0/\underline{40.3} &
  \textbf{98.5}/\textbf{47.5}/\textbf{40.7} \\
BTAD~\cite{BTAD} &
  97.5/54.7/50.6 &
  97.6/51.1/50.0 &
  97.0/50.1/47.5 &
  \underline{97.7}/\underline{57.1}/\underline{56.5} &
  88.7/37.8/32.2 &
  97.5/48.1/46.4 &
  96.0/50.2/39.4 &
  97.5/54.4/51.5 &
  \textbf{97.8}/\textbf{63.6}/\textbf{64.5} \\
DTD-Synthetic~\cite{DTD} &
  91.8/37.6/29.2 &
  95.3/45.1/33.1 &
  96.0/70.6/70.8 &
  97.4/\underline{72.3}/75.5 &
  96.6/52.1/43.3 &
  \underline{98.1}/71.0/\underline{76.8} &
  94.4/61.1/60.9 &
  93.7/61.1/59.3 &
  \textbf{98.8}/\textbf{77.3}/\textbf{80.9} \\ \bottomrule[1.2pt]
\end{tabular}
}
}
\end{table*}
% \usepackage{booktabs}
\begin{table*}[]
\centering
\caption{Quantitative comparisons of VarAD with alternative methods on public datasets. Results are presented as image-level AUROC\%/max-F1\%/AP\%. Best scores are highlighted in \textbf{bold}, while the second-best scores are also \underline{underlined}.}
\label{tab:main_image}
\resizebox{1.\linewidth}{!}{
{
\begin{tabular}{ccccccccc>{\columncolor{blue!8}}c}
\toprule[1.2pt]
Dataset &
\begin{tabular}[c]{@{}c@{}}PFM~\cite{PFM}\\      TII'2022\end{tabular} &
\begin{tabular}[c]{@{}c@{}}RD4AD~\cite{RD4AD}\\      CVPR'2022\end{tabular} &
\begin{tabular}[c]{@{}c@{}}PatchCore~\cite{Patchcore}\\      CVPR'2022\end{tabular} &
\begin{tabular}[c]{@{}c@{}}CDO~\cite{CDO}\\      TII'2023\end{tabular} &
\begin{tabular}[c]{@{}c@{}}PyramidFlow~\cite{PyramidFlow}\\      CVPR'2023\end{tabular} &
\begin{tabular}[c]
{@{}c@{}}MSFlow~\cite{MSFlow}\\      TNNLS'2024\end{tabular} &
\begin{tabular}[c]{@{}c@{}}AMI-Net~\cite{AMI-Net}\\      TASE'2024\end{tabular} &
\begin{tabular}[c]{@{}c@{}}PNPT~\cite{PNPT}\\      TII'2024\end{tabular} &
\begin{tabular}[c]{@{}c@{}}\textbf{VarAD}\\      \textbf{Proposed}\end{tabular} \\ \midrule
MVTec AD~\cite{MVTec-AD} &
  53.8/84.2/75.9 &
  68.3/87.7/81.4 &
  93.1/92.7/96.6 &
  \underline{96.2}/\underline{95.5}/\underline{98.4} &
  80.2/89.0/90.6 &
  \textbf{97.2}/\textbf{95.8}/\textbf{98.6} &
  91.0/92.5/96.0 &
  85.7/90.7/93.1 &
  91.3/92.6/95.8 \\
VisA~\cite{VisA} &
  50.8/72.7/60.5 &
  70.8/83.0/66.8 &
  \underline{96.9}/\underline{93.4}/\underline{97.4} &
  \textbf{97.4}/\textbf{94.2}/\textbf{97.7} &
  82.1/82.2/82.8 &
  93.2/89.1/94.4 &
  88.2/86.9/88.4 &
  92.7/89.2/93.6 &
  90.3/86.1/91.0 \\
BTAD~\cite{BTAD} &
  87.2/88.4/92.6 &
  91.7/90.6/91.1 &
  94.3/91.7/94.8 &
  \underline{94.9}/91.7/\underline{95.1} &
  86.3/81.0/83.2 &
  91.4/\underline{91.8}/94.8 &
  93.8/91.7/94.8 &
  94.7/91.6/94.3 &
  \textbf{95.1}/\textbf{92.0}/\textbf{95.7} \\
DTD-Synthetic~\cite{DTD} &
  59.8/85.5/81.4 &
  66.4/88.7/83.2 &
  92.4/93.0/97.0 &
  \underline{95.9}/95.7/\underline{98.1} &
  94.3/95.4/97.5 &
  95.7/\underline{95.9}/\underline{98.1} &
  90.5/91.4/94.8 &
  82.8/88.7/92.3 &
  \textbf{96.3}/\textbf{96.3}/\textbf{98.5} \\ \bottomrule[1.2pt]
\end{tabular}
}
}
\end{table*}

\subsubsection{Implementation Details}
This study adopts a default resolution of $1024 \times 1024$ for HRIAD. All images undergo resizing to match the dimensions of $1024 \times 1024$, followed by normalization using the mean and standard deviation obtained from the ImageNet dataset. For image tokenization, DINO (ViT-S/14)~\footnote{\href{https://github.com/facebookresearch/dinov2}{https://github.com/facebookresearch/dinov2}} is utilized as the default method. Tokens from the 4th, 8th, and 12th layers are selected to form multi-hierarchy visual tokens, each possessing a channel size of $C=384$. The hyperparameter prediction step $m$ is by default set to 4. The training process utilizes the AdamW optimizer with a learning rate of $5 \times 10^{-4}$ for 10 epochs. All experiments are performed on a computing system equipped with Xeon(R) Gold 6226R CPUs@2.90 GHz, accompanied by one NVIDIA 3090-Ti GPU with 24GB of memory.

\subsubsection{Evaluation Metrics}

{This study employs three widely used metrics to evaluate anomaly detection performance: the Area Under the Receiver Operating Characteristic Curve (AUROC), the maximum F1 Score under optimal thresholds (max-F1), and Average Precision (AP). The primary focus of the evaluation is on pixel-level anomaly detection, while image-level metrics are reported only for the main experiments. Additionally, the Per-Region Overlap (PRO) score is a metric used to assess pixel-level performance in conventional low-resolution settings. However, due to its computational intensity and infeasibility in HRIAD, it is excluded from this study.
}

\subsubsection{Comparison Methods}
{This study conducts a comparative analysis of the proposed VarAD with several popular anomaly detection methods. Specifically, the comparison methods include CNN-based methods, PFM \cite{PFM}, RD4AD \cite{RD4AD}, PatchCore \cite{Patchcore}, CDO \cite{CDO}, PyramidFlow \cite{PyramidFlow}, and MSFlow~\cite{MSFlow}, and transformer-based methods, AMI-Net~\cite{AMI-Net} and PNPT~\cite{PNPT}. 
We utilize their publicly available implementations, only adjusting input resolutions to assess their performance in high-resolution image anomaly detection.}

% in this study to evaluate their performance in high-resolution image anomaly detection.

\begin{figure*}[t]
    \centering
    \includegraphics[width=.95\linewidth]{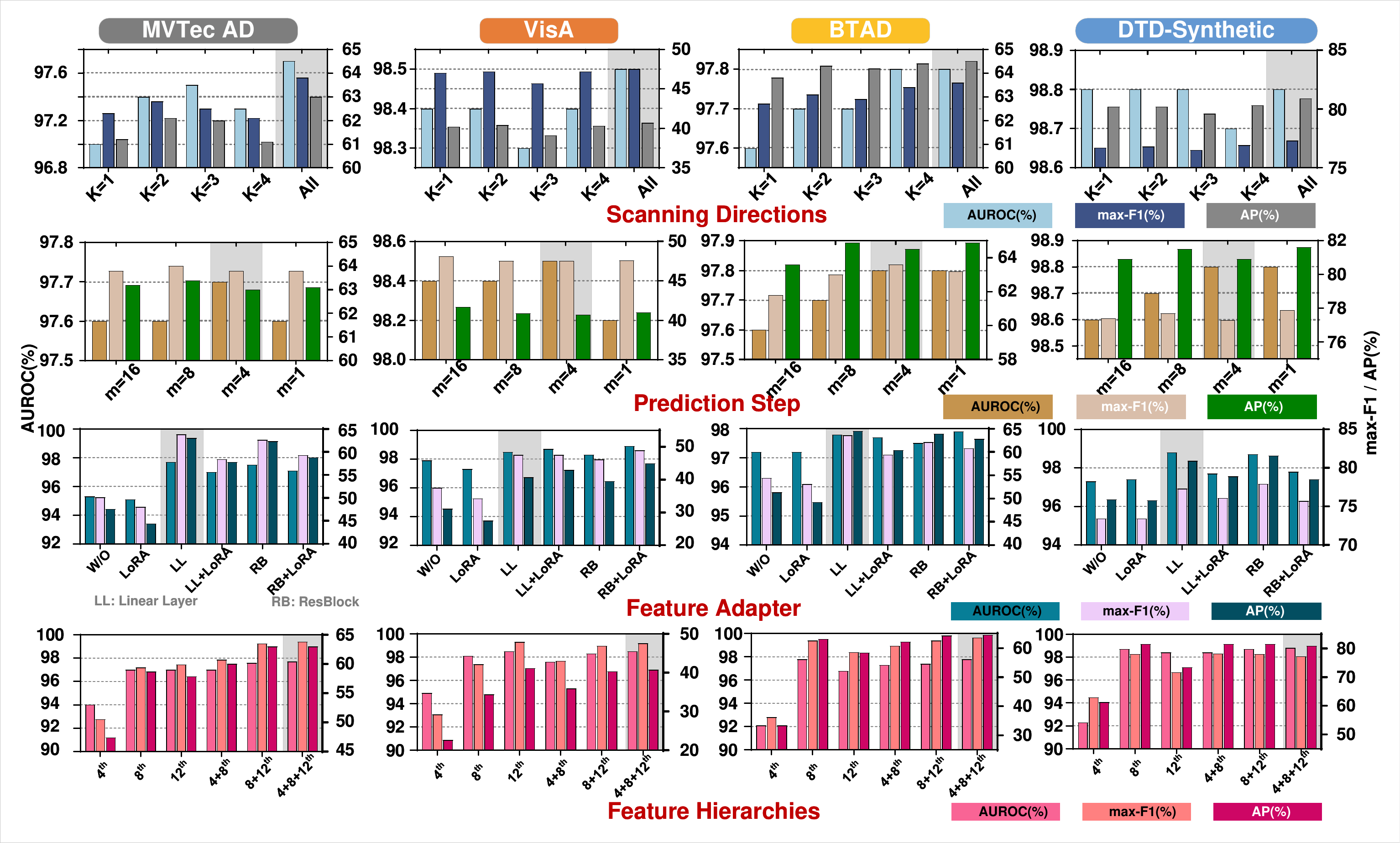}
    \caption{{\textbf{Anomaly detection performance of VarAD with different setups.} From top to bottom: different scanning directions; different prediction steps $m$; different feature adapters, where w/o indicates that the pre-trained model is used to directly extract tokens; and different feature hierarchies.}}
    \label{fig:main_ablation}
\end{figure*}

\subsection{Main Results}

\subsubsection{Pixel-level Comparisons}
Table~\ref{tab:main_pixel} illustrates the qualitative comparisons between VarAD and other alternatives in the proposed HRIAD setting. While representative methods such as PatchCore~\cite{Patchcore} and RD4AD~\cite{RD4AD} have reported saturated performance in their original reports under the low-resolution setting, their detection performance for high-resolution images shows a notable decline. CDO~\cite{CDO} achieves promising detection performance for high-resolution images because of its utilized backbone HRNet~\cite{HRNet}, demonstrating the importance of global information in HRIAD. {MSFlow also achieves promising results thanks to the designed multi-scale flow model.}  Surprisingly, while AMI-Net~\cite{AMI-Net} and PNPT~\cite{PNPT} utilize vision transformers that should be able to capture global information, they still achieve subpar anomaly detection performance in comparison to other alternatives.
In contrast, VarAD achieves the best performance across all datasets, with AUROC scores of 97.7\%, 98.5\%, 97.8\%, and 98.8\% on the four datasets, respectively. VarAD exhibits significant improvements on BTAD and DTD-Synthetic, outperforming the second-place method CDO by 6.5\% and 5.0\% in max-F1, respectively, which underscore the superior global modeling capacity of the proposed VarAD. Fig.~\ref{fig:main_visual} visualizes the detection results of VarAD and the comparison methods, further demonstrating the superiority of VarAD.

\subsubsection{Image-level Comparisons}
{Image-level anomaly detection results are also critical for HRIAD. As shown in Table~\ref{tab:main_image}, all methods exhibit a notable performance decline in HRIAD compared to their original results in low-resolution settings. For instance, PatchCore~\cite{Patchcore} achieves 99.1\% AUROC on MVTec AD in the low-resolution scenario, but only 93.1\% in HRIAD. Overall, MSFlow and CDO demonstrate superior performance on MVTec AD and VisA, whereas our proposed VarAD achieves the highest performance on BTAD and DTD-Synthetic, surpassing the second-place methods by 0.2\% and 0.4\% AUROCs. }

\subsection{Ablation Study}
This study performs a series of ablation experiments to thoroughly assess the impact of individual components within VarAD. Given the primary focus on pixel-level performance in this study, only pixel-level metrics are reported in the following experiments.

\begin{table*}[t]
\centering
\caption{Ablation study on different backbones for VarAD.}
\label{tab:backbone}
\setlength\tabcolsep{12.0pt}
\resizebox{1.\linewidth}{!}{
\begin{tabular}{cccc>{\columncolor{blue!8}}ccc}
\toprule[1.2pt]
Backbone  & ViT-B/16 & MAE-ViT-Base                  & CLIP-ViT-B/16      & DINO-ViT-S/14      & DINO-ViT-B/14      & DINO-ViT-L/14      \\ \midrule
MVTec AD~\cite{MVTec-AD}& 
93.6/51.8/48.9 & 
61.5/12.5/07.7 & 
95.5/53.5/51.6 & 
\underline{97.7}/\underline{63.8}/\underline{63.0} &
\textbf{97.9}/\textbf{65.1}/\textbf{64.6} & 
95.6/55.2/52.5 \\
VisA~\cite{VisA}      & 
90.9/30.6/24.5 & 
59.2/03.5/01.4  & 
95.5/35.5/27.6 & 
\underline{98.5}/\underline{47.5}/\underline{40.7} & 
\textbf{98.9}/\textbf{52.9}/\textbf{48.7} & 96.7/38.4/32.4 \\
BTAD~\cite{BTAD}     & 
91.3/32.4/29.3  & 
72.6/14.3/08.2  & 96.0/49.8/49.3 & 
\underline{97.8}/\underline{63.6}/\underline{64.5} & 
\textbf{98.1}/\textbf{66.2}/\textbf{68.3} & 95.0/39.6/36.1 \\
DTD-Synthetic~\cite{DTD}      & 
88.4/60.2/58.1  & 
53.5/02.7/01.3  & 
90.8/61.7/60.8 & 
\underline{98.8}/\underline{77.3}/\underline{80.9} &
\textbf{99.1}/\textbf{78.5}/\textbf{82.7} & 
95.3/71.7/72.5 \\ \midrule
Params (M)/FPS & 112.8/2.9  & 112.2/7.4                           & 112.8/2.9              & {22.7}/8.0               & 88.2/3.8               & 335.1/1.7 
\\
% FPS       & 2.9   & 7.4                              & 2.9                & {8.0}                & 3.8                & 1.7                \\ 
\bottomrule[1.2pt]
\end{tabular}
}

\end{table*}

\subsubsection{Influence of Sequentialized Direction}

During the token sequentialization process, VarAD scans the visual tokens in multiple directions to prevent potential information loss. Fig.~\ref{fig:main_ablation} presents a comparison of the detection performance of VarAD when scanning in different directions. The results indicate that aggregating information from all directions consistently yields superior performance compared to using a single direction. This underscores the effectiveness of utilizing multi-directional visual token sequences.

\subsubsection{Hyperparameter Sensitive Analysis}
The only hyperparameter for VarAD is $m$, which controls the length of neighboring tokens to be excluded in the token prediction process. This study assesses the detection performance of VarAD with different values of $m$, and the results are presented in Fig.~\ref{fig:main_ablation}. Generally, the prediction process becomes more challenging with increased $m$, as more information from neighboring tokens is excluded. Hence, with increased $m$, VarAD is expected to acquire a stronger global modeling capacity. However, a larger $m$ may lead to the loss of local information and influence the prediction quality. As visualized in Fig.~\ref{fig:main_ablation}, the detection performance exhibits an increasing and then decreasing trend with larger values of $m$, peaking at $m=4$. Nevertheless, VarAD is not sensitive to changes in $m$, showing only slight variations in detection performance under different values of $m$.

% This is attributed to the fact that $m=4$ can capture both significant local and global information compared to other values. 

\subsubsection{Influence of Feature Adapter}

The visual token sequence extraction process involves tokenizing images into visual tokens. VarAD defaults to utilizing the pre-trained model DINO for tokenization and proposes a feature adapter to mitigate the domain gap between the pre-trained natural image data and the target industrial data. We design several feature adapters for evaluation, including a simple linear layer and a more complex ResBlock. In addition to appending the feature adapters after the pre-trained model, we also explore a parameter-efficient fine-tuning method, LoRA~\cite{Lora}, and its combination with the designed feature adapters, as shown in Fig.~\ref{fig:main_ablation}.
It is evident from the table that the feature adapter, whether a linear layer or a ResBlock, significantly improves detection performance, while solely using LoRA fails to mitigate the domain gap and leads to subpar detection results. Combining LoRA with the linear layer or ResBlock shows improvements in certain metrics but declines in others. Furthermore, the complex adapter ResBlock does not consistently outperform the simple linear layer but can introduce more parameters. Therefore, we default to using only the linear layer for feature adaptation, improving more than 10.0\% in max-F1 on MVTec AD, VisA, and BTAD compared to the unadapted model.

\subsubsection{Influence of Token Hierarchy}
This study evaluates VarAD under different combinations of token hierarchies. Generally, tokens from a shallow layer contain structural information, while those from a deeper layer may incorporate more global information. As shown in Fig.~\ref{fig:main_ablation}, the combination of the 4th, 8th, and 12th token hierarchies consistently achieves better performance compared to other combinations. Therefore, this combination is selected as the default setting for VarAD.

\subsubsection{Influence of Backbone}

To comprehensively investigate the influence of backbones, this study compares VarAD with different backbones, including ViT~\cite{ViT} undergoing supervised pre-training, MAE~\cite{MAE}, CLIP~\cite{clip}, and DINO with various architectures. The comparison results are shown in Table~\ref{tab:backbone}. It clearly demonstrates that pre-trained backbones significantly influence detection performance, indicating that backbones trained with specific objectives may better suit visual autoregressive modeling. We will further study the underlying mechanisms in the future. Overall, VarAD with DINO-ViT-B/14 achieves the best anomaly detection performance. However, DINO-ViT-B/14 requires nearly four times the parameters compared to DINO-ViT-S/14 and achieves only 3.8 FPS. Consequently, after trading off all factors, this study elects to use DINO-ViT-S/14 as the visual tokenizer, achieving lightweight, efficient, and promising high-resolution anomaly detection performance.

\begin{figure}[t]
    \centering
    \includegraphics[width=1\linewidth]{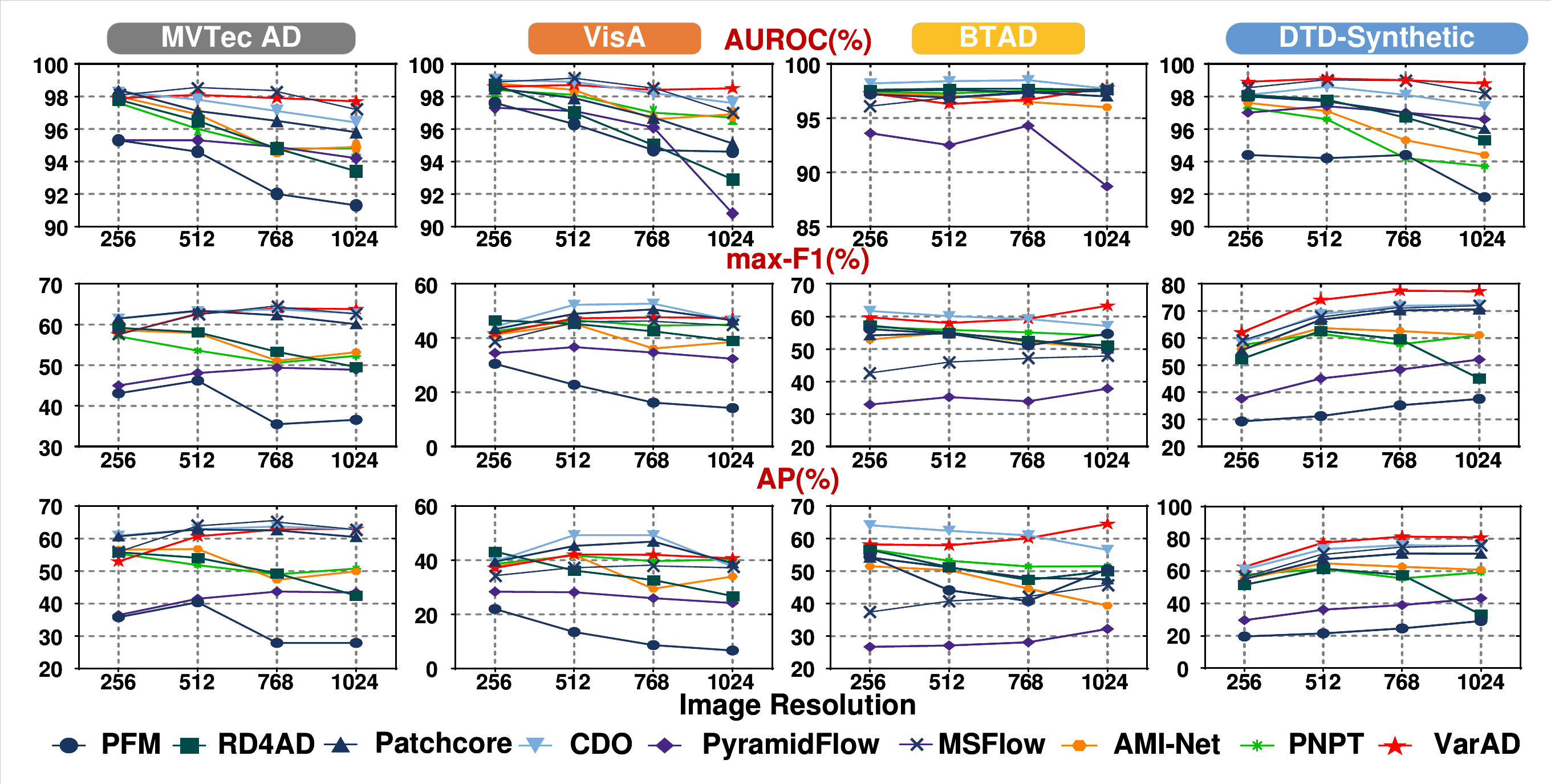}
    \caption{{Anomaly detection performance under different resolutions.}}
    \label{fig:res}
\end{figure}

\subsection{Analysis}

\subsubsection{Influence of Image Resolution}
\begin{figure}[t]
    \centering
    \includegraphics[width=1\linewidth]{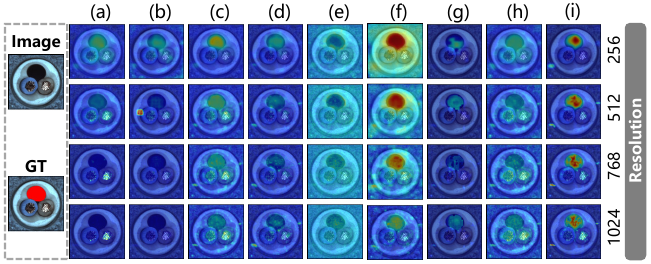}
    \caption{\textbf{Anomaly maps under different resolutions.} (a) PFM, (b) RD4AD, (c) PatchCore, (d) CDO, (e) PyramidFlow, (f) MSFlow, (g) AMI-Net, (h) PNPT, (i) VarAD.}
    \label{fig:res_visual}
\end{figure}

This study evaluates the detection performance of comparison methods under different resolutions, namely $256 \times 256$, $512 \times 512$, $768 \times 768$, $1024 \times 1024$. Fig.~\ref{fig:res} presents their detection performance on the four datasets under different resolutions. It shows that the existing AD methods tend to perform weaker with larger resolutions. For instance, the AUROC on MVTec AD of PatchCore decreases from 98.4\% to 95.8\% when image resolutions increase from $256 \times 256$ to $1024 \times 1024$. While CDO can better capture global information compared to other existing AD methods, it still witnesses some drops in detection performance. In contrast, the proposed VarAD achieves commendable detection performance across all resolutions. Its AUROC remains stable for different resolutions. VarAD even witnesses some slight improvements in max-F1 and AP with increased resolutions for all datasets. Whereas VarAD may underperform compared to other methods at low resolutions, such as $512\times512$ on VisA, certain scenarios necessitate high-resolution images to address subtle anomalies. In these cases, subtle anomalies may occupy very few pixels (\eg, five) in low-resolution images, making them difficult to detect. 
Fig.~\ref{fig:res_visual} further visualizes the detection results of these comparison methods under different resolutions. It is evident that VarAD consistently achieves excellent detection results under improved resolutions, while other methods exhibit deficiencies. 
% This effectively demonstrates the superior detection performance of VarAD under different resolutions, presenting a viable solution for practical applications.

\begin{figure}[t]
    \centering
    \includegraphics[width=1\linewidth]{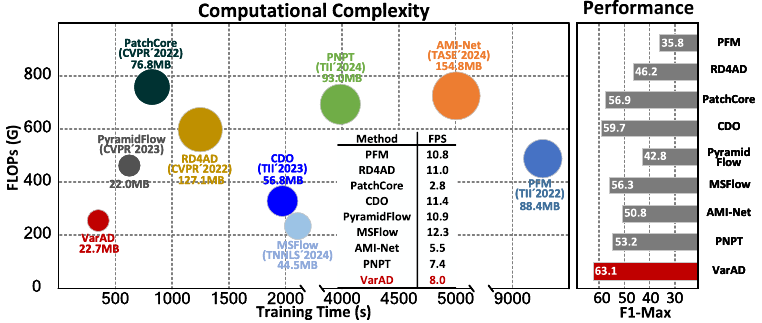}
    \caption{Complexity comparisons of different methods.}
    \label{fig:computational_complexity}
\end{figure}
\subsubsection{Complexity Analysis}
Considering practical applications, both anomaly detection performance and model efficiency require comprehensive evaluation. Specifically, this study conducts a fair comparison of various methods across four dimensions: 1) training time—measured on a single NVIDIA 3090Ti with a batch size of two, 2) number of parameters, 3) floating-point operations per second (FLOPs), and 4) inference speed in terms of frames per second (FPS). Fig.~\ref{fig:computational_complexity} demonstrates that transformer-based methods exhibit higher computational complexity compared to others, while CNN-based methods generally offer greater efficiency.
Overall, the proposed VarAD achieves the fastest training time, significantly lower FLOPs, and the highest average max-F1 score across the four datasets. Specifically, VarAD converges within only 10 epochs (389s), whereas other methods, such as CDO and RD4AD, typically require 50 epochs (over 1200s). {Moreover, the parameter count and FLOPs of VarAD are lower than those of other methods, with only a slight increase compared to PyramidFlow and MSFlow, yet demonstrating significantly better performance. In terms of inference speed, VarAD operates in an autoregressive manner, leading to a slightly lower FPS than CNN-based methods such as PyramidFlow and MSFlow, which utilize efficient flow models for rapid inference. Nevertheless, VarAD achieves a commendable speed of 8.0 FPS.
}

\begin{table*}[]
\centering
\caption{Quantitative comparisons of VarAD with alternative methods on the collected button inspection dataset. Results are presented as AUROC\%/max-F1\%/AP\%. Best scores are highlighted in \textbf{bold}, while the second-best scores are \underline{underlined}.}
\label{tab:button}
\resizebox{1.\linewidth}{!}{
\begin{tabular}{ccccccccc>{\columncolor{blue!8}}c}
\toprule[1.2pt]
Dataset &
\begin{tabular}[c]{@{}c@{}}PFM~\cite{PFM}\\      TII'2022\end{tabular} &
\begin{tabular}[c]{@{}c@{}}RD4AD~\cite{RD4AD}\\      CVPR'2022\end{tabular} &
\begin{tabular}[c]{@{}c@{}}PatchCore~\cite{Patchcore}\\      CVPR'2022\end{tabular} &
\begin{tabular}[c]{@{}c@{}}CDO~\cite{CDO}\\      TII'2023\end{tabular} &
\begin{tabular}[c]{@{}c@{}}PyramidFlow~\cite{PyramidFlow}\\      CVPR'2023\end{tabular} &
\begin{tabular}[c]
{@{}c@{}}MSFlow~\cite{MSFlow}\\      TNNLS'2024\end{tabular} &
\begin{tabular}[c]{@{}c@{}}AMI-Net~\cite{AMI-Net}\\      TASE'2024\end{tabular} &
\begin{tabular}[c]{@{}c@{}}PNPT~\cite{PNPT}\\      TII'2024\end{tabular} &
\begin{tabular}[c]{@{}c@{}}\textbf{VarAD}\\      \textbf{Proposed}\end{tabular} \\ \midrule
Class1  & 
95.1/9.2/3.4 & 
96.2/\underline{30.2}/15.4 & 
95.1/27.5/14.2 & 
95.9/24.6/\underline{19.1} & 
\underline{97.0}/23.7/13.3 & 
96.4/10.3/4.6  & 
94.1/27.6/14.5 & 
96.2/24.8/17.3 & 
\textbf{97.1}/\textbf{41.8}/\textbf{38.7} \\
Class2  & 
92.6/3.8/1.6 & 
95.9/25.9/11.6 & 
93.4/19.5/12.6 & 
\underline{97.6}/\underline{39.7}/\textbf{37.7} & 
93.7/14.3/8.0 & 
93.7/11.3/8.5  & 
92.8/34.9/30.1 & 
94.3/37.4/31.4 & 
\textbf{98.0}/\textbf{42.3}/\underline{35.3} \\
Class3  & 
88.6/2.8/1.1 & 
\underline{96.8}/23.3/12.0 & 
91.2/18.1/11.2 & 
\textbf{98.3}/\textbf{38.4}/\textbf{35.0} & 
88.8/8.8/3.7  & 
96.3/25.7/\underline{24.1} & 
91.3/26.1/19.6 & 
89.9/27.1/22.3 & 
94.8/\underline{31.7}/24.0 \\
Class4  & 
90.9/6.3/1.5 & 
\underline{97.2}/21.0/10.0 & 
91.7/25.9/\underline{16.8} & 
95.0/19.8/13.0 & 
93.5/8.3/3.8  & 
94.4/18.2/12.5 & 
94.2/\underline{27.0}/11.4 & 
94.1/24.3/15.7 & 
\textbf{97.3}/\textbf{28.1}/\textbf{25.2} \\ \midrule
Average & 
91.8/5.5/1.9 & 
96.5/25.1/12.3 & 
92.8/22.8/13.7 & 
\underline{96.7}/\underline{30.6}/\underline{26.2} & 
93.2/13.8/7.2  & 
95.2/16.4/12.4 & 
93.1/28.9/18.9 & 
93.6/28.4/21.7 & 
\textbf{96.8}/\textbf{36.0}/\textbf{30.8} \\ \bottomrule[1.2pt]
\end{tabular}
}
\end{table*}

\begin{figure}[t]
    \centering
    \includegraphics[width=.8\linewidth]{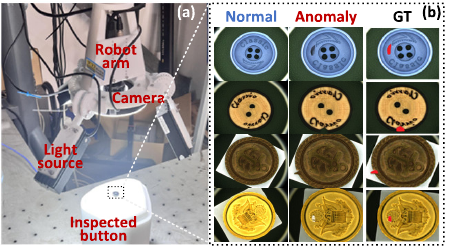}
    \caption{(a) Real-world button inspection platform. (b) Samples.}
    \label{fig:real_world_platform}
\end{figure}

% \usepackage{booktabs}
% \usepackage{multirow}
\begin{table}[t]
\centering
\caption{Statistical information of the collected button dataset. \# denotes the number of samples.}
\label{tab:button_dataset_info}
\resizebox{1\linewidth}{!}{
{
\begin{tabular}{@{}cc|cccc|c@{}}
\toprule[1.2pt]
\multicolumn{2}{c|}{Dataset}           & Class1 & Class2 & Class3 & Class4 & Summary \\ \midrule
Training                 & \#Normal   & 70     & 70 & 40 & 50          & 230   \\
\multirow{2}{*}{Testing} & \#Normal   & 70      & 70  & 40  & 50           & 230    \\
                         & \#Abnormal & 26     & 26 & 20  & 15           & 87    \\ \bottomrule[1.2pt]
\end{tabular}
}
}
\end{table}

\subsection{Evaluations on Real-World Data}

To further assess the applicability and effectiveness of VarAD, this study develops an image acquisition platform, as depicted in Fig.~\ref{fig:real_world_platform} (a), and gathers private data from a real-world button inspection task.
During the image acquisition process, various types of noise are intentionally introduced to assess the robustness of the comparison methods, including unaligned positions of buttons, different illumination conditions, and background interference. These noise types render the established real-world button inspection dataset more challenging than the four public datasets utilized in this study. Additionally, these buttons feature potential subtle and small anomalies, necessitating high-resolution image anomaly detection.  Hence, we acquire images with a high resolution of $1024\times1024$. Representative samples from the dataset are visualized in Fig.~\ref{fig:real_world_platform} (b). The constructed dataset consists of four categories of buttons. Table~\ref{tab:button_dataset_info} shows the detailed statistical information for the collected button datasets.

Table~\ref{tab:button} provides a category-level quantitative comparison of results on the established dataset. It is apparent that all methods demonstrate weaker performance on this dataset compared to their results on public datasets, indicating the increased difficulty of the established dataset. Nevertheless, VarAD achieves notably superior detection results compared to other methods, with impressive scores of 96.8\% AUROC, 36.0\% max-F1, and 30.8\% AP. VarAD surpasses the second-place method, CDO, by 0.1\% AUROC, 5.4\% max-F1, and 4.6\% AP. The qualitative anomaly localization comparison results of various methods are depicted in Fig.~\ref{fig:main_visual} (Button). It is evident that the proposed VarAD achieves significantly better detection results than other alternatives.

\section{Conclusion}

In summary, this paper presents VarAD, a novel approach for detecting anomalies in high-resolution images. VarAD transforms anomaly detection into a token prediction task, utilizing Mamba for visual autoregressive modeling. By predicting future tokens in multi-hierarchy and multi-directional visual token sequences, VarAD effectively scores anomalies based on prediction errors. Extensive evaluations on four publicly available datasets and a real-world button inspection dataset attest to the superior detection performance and efficiency of VarAD. {In future research, we aim to extend VarAD to simultaneous anomaly detection across multiple categories and improve its image-level anomaly detection performance.}

\bibliographystyle{ieeetr} 
% \bibliography{reference}

\vspace{-8mm}
\begin{IEEEbiography}[{\includegraphics[width=1in,height=1.25in,clip,keepaspectratio]{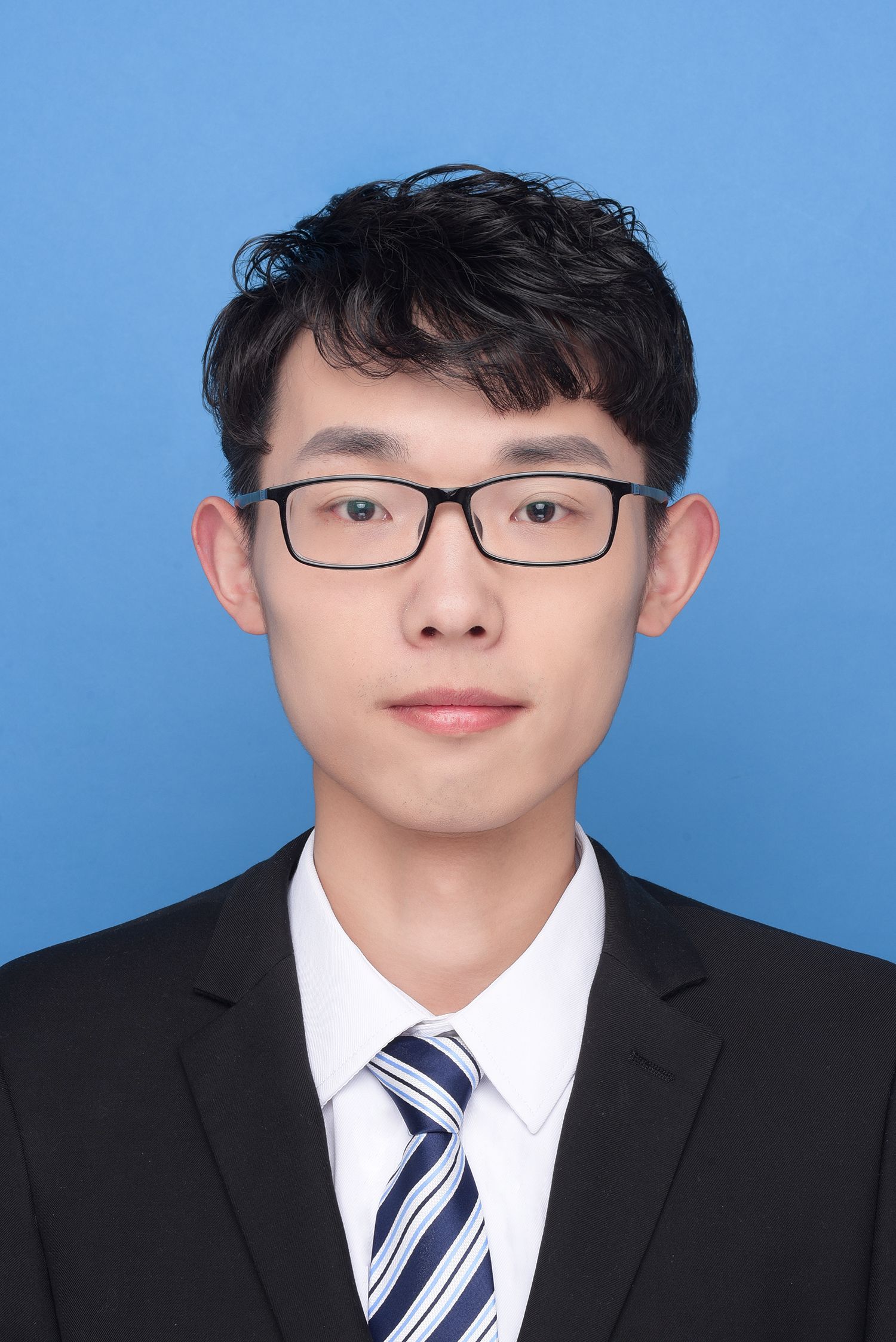}}]{Yunkang Cao} (Graduate Student Member, IEEE) received his B.S. degree from Huazhong University of Science and Technology (HUST), Wuhan, China, in 2020, where he is currently pursuing his Ph.D. degree in mechanical engineering. 
His current research interests include machine vision, visual anomaly detection, and industrial foundation models. 
\end{IEEEbiography}

\vspace{-8mm}
\begin{IEEEbiography}[{\includegraphics[width=1in,height=1.25in,clip,keepaspectratio]{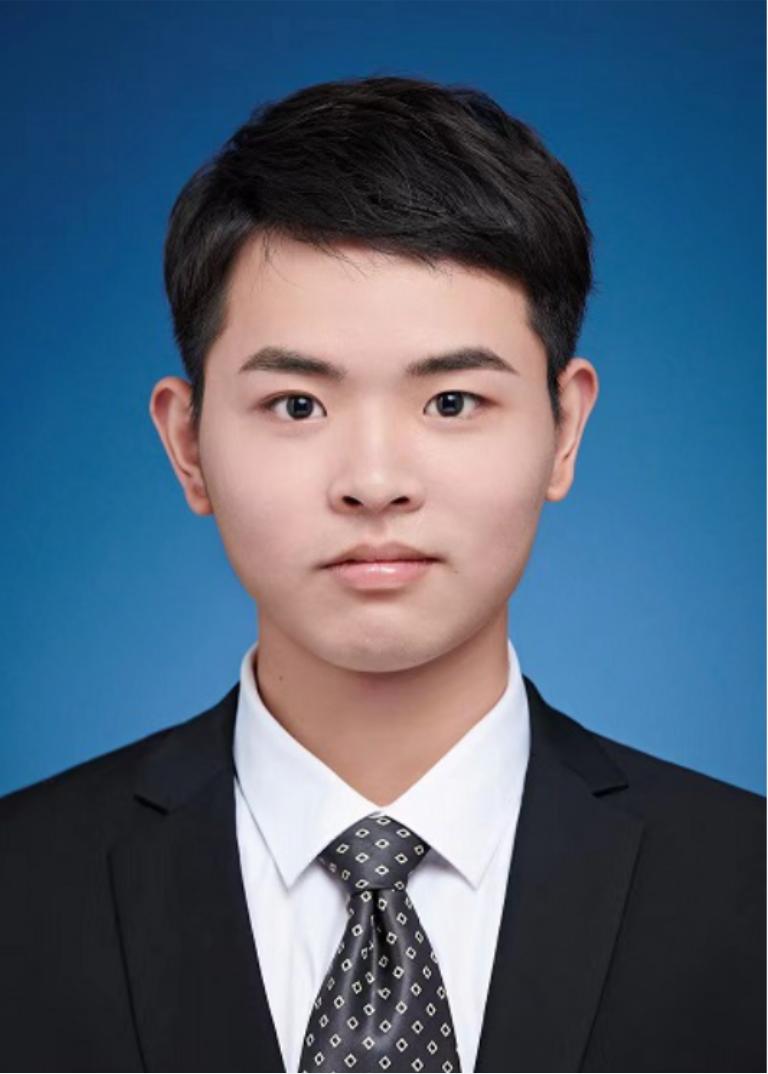}}]{Haiming Yao}  (Graduate Student Member, IEEE) received a B.S. degree (Hons.) from the School of Mechanical Science and Engineering, Huazhong University of Science and Technology, Wuhan, China, in 2022. He is pursuing a Ph.D. degree with the Department of Precision Instrument, Tsinghua University, Beijing, China. His research interests include visual anomaly detection, deep learning, visual understanding, and artificial intelligence for science.
\end{IEEEbiography}

\vspace{-8mm}
\begin{IEEEbiography}[{\includegraphics[width=1in,height=1.25in,clip,keepaspectratio]{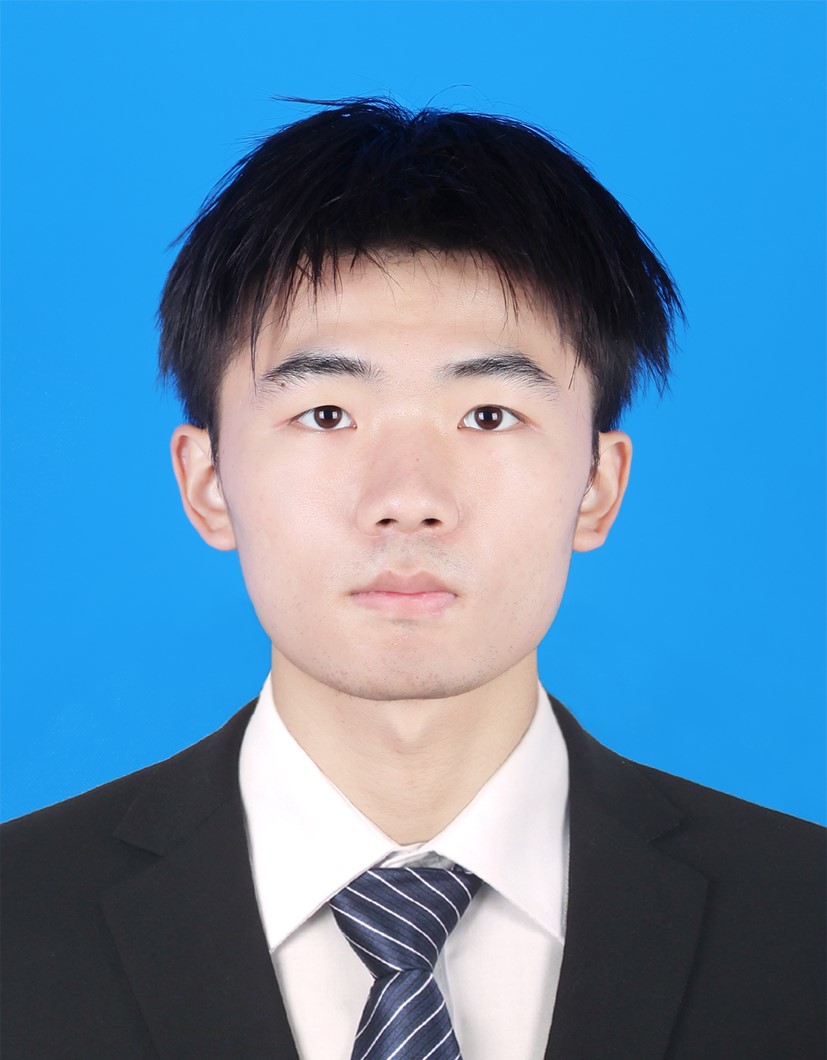}}]{Wei Luo}
(Student Member, IEEE) received a B.S. degree from the School of Mechanical Science and Engineering, Huazhong University of Science and Technology, Wuhan, China, in 2023. He is pursuing a Ph.D. degree with the Department of Precision Instrument, Tsinghua University, Beijing, China. His research interests include deep learning, anomaly detection, and machine vision.
\end{IEEEbiography}

\vspace{-8mm}
\begin{IEEEbiography}[{\includegraphics[width=1in,height=1.25in,clip,keepaspectratio]{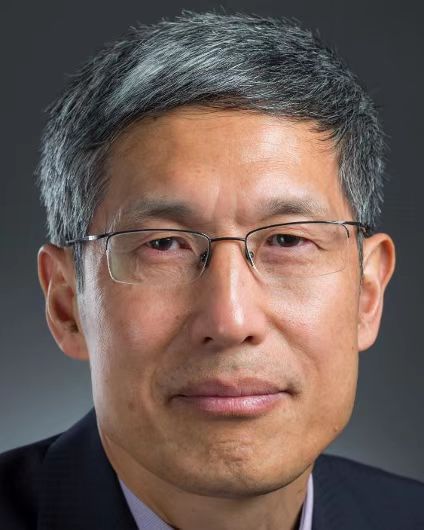}}]{Weiming Shen} (Fellow, IEEE) received the B.E. and M.S. degrees in mechanical engineering from Northern Jiaotong University, Beijing, China, in 1983 and 1986, respectively, and the Ph.D. degree in system control from the University of Technology of Compiegne, Compiegne, France, in 1996. He is currently a Professor at the Huazhong University of Science and Technology (HUST), Wuhan, China, and an Adjunct Professor at the University of Western Ontario, London, ON, Canada. Before joining HUST in 2019, he was a Principal Research Officer at the National Research Council Canada. He is a Fellow of Canadian Academy of Engineering and the Engineering Institute of Canada. 

His work has been cited more than 16,000 times with an h-index of 61. He authored or co-authored several books and more than 560 articles in scientific journals and international conferences in related areas. His research interests include agent-based collaboration technologies and applications, collaborative intelligent manufacturing, the Internet of Things, and Big Data Analytics.

\end{IEEEbiography}

\end{document}